%% file: main.tex
\def\year{2021}\relax
\documentclass[letterpaper]{article} 

\usepackage{aaai21}  
\usepackage{times}  
\usepackage{helvet} 
\usepackage{courier}  
\usepackage[hyphens]{url}  
\usepackage{graphicx} 
\def\UrlFont{\rm}  
\usepackage{natbib}  
\usepackage{caption} 
\usepackage{graphicx}
\usepackage{amsmath}
\usepackage{amsthm}
\usepackage{booktabs}
\usepackage{enumitem}

\newtheorem{theorem}{Theorem}
\newtheorem{definition}{Definition}

\usepackage{latexsym}
\usepackage{subcaption}
\usepackage{xcolor}
\usepackage[switch]{lineno}

\newcommand{\MikeNote}[1]{\textcolor{red}{Mike: #1}}

\newcommand{\pjs}[1]{\textcolor{purple}{PJS: #1}}
\newcommand{\ignore}[1]{}
\title{Symmetry Breaking for $k$-Robust Multi-Agent Path Finding}

\begin{document}
\author{Zhe Chen,\textsuperscript{\rm 1} 
        Daniel Harabor,\textsuperscript{\rm 1} 
        Jiaoyang Li,\textsuperscript{\rm 2}\protect\thanks{Jiaoyang Li performed the research during her visit to Monash University.} 
        Peter J. Stuckey\textsuperscript{\rm 1} \\}
\affiliations{\textsuperscript{\rm 1}Monash University, Australia \\
             \textsuperscript{\rm 2}University of Southern California, USA \\
             \{zhe.chen,daniel.harabor,peter.stuckey\}@monash.edu, jiaoyanl@usc.edu}

\maketitle

\input abstract
\input introduction
\input background

\input k_cbs
\input symmetries

\input reasoning

\input k_Corridor_Target

\input figures
\input experiments

\input conclusion

\bibliography{reference}

\end{document}

%% file: abstract.tex
\begin{abstract}
During Multi-Agent Path Finding (MAPF) problems, agents can be delayed
by unexpected events. To address such situations recent work describes
$k$-Robust Conflict-Based Search ($k$-CBS): an algorithm that
produces a coordinated and collision-free plan that is robust for up to $k$ delays for any agent. 
In this work we introduce a variety of pairwise 
symmetry breaking constraints, specific to $k$-robust planning, that can efficiently find
compatible and optimal paths for pairs of colliding agents. 
We give a thorough description of the new constraints and report
large improvements to success rate in a range of domains including:
(i) classic MAPF benchmarks, (ii) automated warehouse domains, and 
(iii) on maps from the 2019 Flatland Challenge, a recently introduced railway
domain where $k$-robust planning can be fruitfully applied to schedule trains.
\end{abstract}

%% file: introduction.tex
\section{Introduction}
Multi-Agent Path Finding (MAPF) is a coordination problem
where we need to find collision-free paths for a team of
cooperating agents and is known to be NP-hard on graphs and grids \cite{yu2013structure,banfi2017intractability}.
When MAPF problems are solved in practice, agents can 
sometimes be unexpectedly delayed during plan execution; e.g. due 
to exogenous events or mechanical problems.
Currently, there exist two principal approaches to handle such delays.  
The first approach involves {\em robust execution policies}~\cite{ma2017multi,honig2019}.
Here dependencies are introduced to guarantee that agents execute their 
plans in a specific and compatible order.
Another approach is to reason about potential delays at the planning stage
which involves computing {\em robust plans}.

Following~\cite{atzmon} we say that a plan is $k$-robust if the individual 
path of each agent remains valid for up to $k$ unexpected delays of that 
agent. In other words, provided each agent waits for no more than $k$ timesteps
on the way to its target, its plan is guaranteed to be collision-free. 
In addition to their execution benefits, $k$-robust plans are valuable in
application areas where agents must maintain minimum safety distances.
Such constraints appear in rail scheduling, quay crane scheduling, planning for
warehouse robots and others. 
In these settings a $k$-robust plan naturally provides $k$ timesteps of 
distance between agents that are are moving, while also allowing agents to stay 
close if they are not moving. Furthermore, $k$-robust plans can be used in conjunction with \emph{robust execution policies} to benefit from both methods \cite{atzmon2020robust}.

To compute $k$-robust plans, \citet{atzmon} propose $k$-robust Conflict-Based
Search ($k$-CBS), a robust variant of the popular and well known branch-and-replan
strategy used for MAPF (where $k$ = 0)~\cite{sharon}, and a SAT-based solution, which cannot solve problems on large grids like brc202d DAO map\cite{Stern2019}.
A main problem with CBS is that the algorithm is extremely inefficient
when reasoning \emph{equivalent permutations} of conflicts that
can occur between pairs of agents, such as rectangle symmetry~\cite{li2019}, 
corridor symmetry, and target symmetry~\cite{corridor}.
Further complicating the situation is that the reasoning techniques proposed
to handle these situations, do not always extend straightforwardly to the $k$-robust case.
\ignore{
especially those for {\em rectangle symmetries}, cannot
be easily extended to computing $k$-robust plans. 
}
\ignore{
In Section \ref{section:sym}, the paper illustrates $k$-robust rectangle
conflicts and why existing rectangle reasoning methods no longer work.
}

To address this gap in the literature we introduce a variety of specialised 
$k$-robust symmetry breaking constraints that dramatically improve performance
for the $k$-CBS algorithm.  Experimental results show very large gains
in success rate for $k$-CBS; not only on classic MAPF benchmarks but
also in two application specific domains: in \emph{warehouse logistics},
where $k$-robust plans are desirable and in \emph{railway scheduling}
where $k$-robust plans are mandatory.

\ignore{
\MikeNote{Talk about the situation that k-robust make rectangle more difficult. \textbf{Briefly here, more in section 4}\\
Agents get out and back. \textbf{In section 5}\\
Talk about,  safety distance. \textbf{Tick} \\
Example of k=4 which makes the problem much complicated and much more flex. \textbf{In Section 5}}
}

%% file: background.tex
\section{Problem Definition}

We consider a multi-agent coordination problem where the operating
environment is an undirected (e.g. gridmap) or directed (e.g. rail network) 
graph $G=(V, E)$. We restrict graph $G$ to be a 4-neighbour grid and we
place upon it $m$ agents $\{a_{1}...a_{m}\}$. 
Every agent $a_i$ is assigned an initial vertex
$s_i$ and a goal vertex $g_i$.  
Time is discretised into unit-size steps. In each timestep, agents can move 
to an adjacent vertex or wait at the current location. Each move or wait
action has an associated unit cost.

We say that a path is $k$-robust if for each location $v$ visited at time $t$ by agent $a$ no other agent
visits the location in the time interval $[t, t+k]$.
We call a $k$-delay \emph{vertex conflict} the situation where agent
$a_i$ at timestep $t$ and $a_j$ at timestep $t'=t+\Delta, \Delta \in [0, k]$, 
visit the same location $v$. We denote such a conflicts 
by $\langle a_i, a_j, v, t, \Delta\rangle$.
When $k=0$ it is further possible for two agents to cross the same edge in opposite directions at the same time, 
resulting in a so-called {\em edge conflict}. Notice however that with $k >0$ such a crossing will always 
result in a vertex conflict. Thus for $k>0$ we do not need to model edge conflicts.

A solution to the problem 
(equiv. a \emph{$k$-robust plan}) is a set of  
paths, one for each agent $a_i$, which moves each agent $a_i$ from its start location $s_i$ to its goal location 
$g_i$ such that there are no $k$-delay conflicts with the plans of any other agent.
Our objective is to find a $k$-robust plan which minimises the 
sum of all individual path costs (SIC).

%% file: k_cbs.tex
\section{$k$-Robust Conflict-Based Search}\label{sec:kCBS}
\begin{figure}
\centering
\begin{subfigure}{0.4\columnwidth}
  \centering
  \includegraphics[width=1\columnwidth]{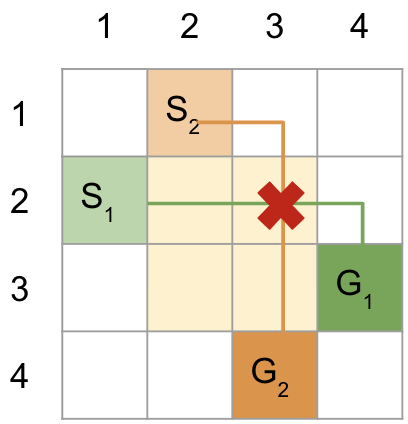}
  \caption{Rectangle}
  \label{fig2:sub0}
\end{subfigure}
\begin{subfigure}{0.43\columnwidth}
  \centering
  \includegraphics[width=1\columnwidth]{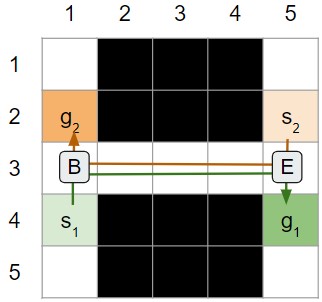}
  \caption{Corridor}
  \label{fig2:sub1}
\end{subfigure}
\begin{subfigure}{0.5\columnwidth}
  \centering
  \includegraphics[width=1\columnwidth]{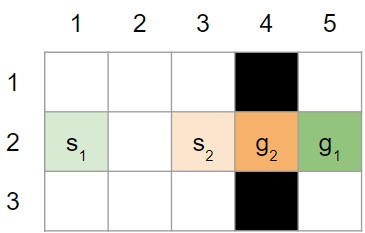}
  \caption{Target}
  \label{fig2:sub2}
\end{subfigure}
\caption{Examples of rectangle, corridor and target conflicts when $k$=0, reproduced from~\cite{corridor}. }
\label{fig2}
\end{figure}

$k$-Robust Conflict-Based Search ($k$-CBS)~\cite{atzmon} is a two-level search 
algorithm specialised from classical MAPF ($k$=0)~\cite{sharon}. 
At the high-level, $k$-CBS searches (in a best-first way) a binary
constraint tree ($CT$) where each  node is a complete assignment of paths to agents (i.e. a plan).  
The process of finding such single agent paths constitutes the low level 
of $k$-CBS. Here individual agents find paths (via A*) from 
start to target while subject to a set of collision-avoiding 
\emph{constraints}. 

The search process of $k$-CBS proceeds as follows:
At each iteration $k$-CBS expands the CT node with lowest $f$-cost. 
If the current node is conflict-free then that node is a goal and 
the search terminates having found a least-cost feasible plan.  
Otherwise, the current node must contain at least one pair of agents 
that are in collision.
Suppose for example that the conflict is $\langle a_i, a_j, v, t, \Delta\rangle$.
%
%
This situation occurs when agent $a_i$ at timestep $t$ and $a_j$ at
timestep $t' = t+\Delta, \Delta \in [0, k]$, visit the same location $v$.
To resolve the conflict $k$-CBS re-plans each of the two affected agents and 
thus generates two new candidate plans. To each child node is added a 
\emph{time range constraint} which resolves the situation now and in 
all future descendant nodes derived from each respective child.
The constraint \mbox{$\langle a_i, v, [t,t+k]\rangle$} says that
agent $a_i$ is not allowed to occupy vertex $v$ at any timestep in the 
range $[t,t+k]$.  The second child node, where $a_j$ is replanned, 
receives a similar constraint: $\langle a_j, v, [t,t+k]\rangle$.

$k$-CBS continues in this way, splitting and searching, 
while the current CT node contains any conflict.
This approach is solution complete and optimal. It guarantees to 
find a $k$-robust plans~\cite{atzmon}, if any such plan exists,
since the union of valid plans permitted by the two child 
nodes is the same as at the parent node
(i.e. adding constraints does not eliminate valid solutions).

\textbf{Conflict selection strategies: }
Deciding which conflict to resolve next is critical to the success 
of ($k$-)CBS. 
In this work we follow \citet{boyarski2015} where authors classify conflicts as 
\emph{cardinal}, \emph{semi-cardinal} and \emph{non-cardinal}:
\begin{itemize}
    \item A conflict C is \emph{cardinal} if replanning for any agent involved in the conflict increases the SIC.
    \item A conflict C is \emph{semi-cardinal} if replanning for one agent involved in the conflict always increases the SIC while replanning for the other agent does not.
    \item A conflict C is \emph{non-cardinal} otherwise.
\end{itemize}
In \citet{boyarski2015} it is shown that resolving cardinal conflicts first can dramatically reduce the size of the resulting CBS tree.
After all cardinal conflicts are resolved we choose semi-cardinal conflicts and finally non-cardinal.
Similar to that work we use a Multi-valued Decision Diagram (MDD) to classify conflicts. 
Each MDD records all nodes (i.e., vertex-time pairs) that can appear on an optimal path 
for each agent. (Semi-)cardinal conflicts require the conflict node to be a \emph{singleton} 
for (resp. one) both agents, i.e. all optimal paths must pass through the node. 

\textbf{CBS Heuristics (CBSH): }
Different from~\cite{atzmon} but following~\cite{CBSH}, we also exploit known cardinal conflicts 
to derive a \emph{minimal cost increase} heuristic for our high-level A* search. This strategy is known to 
improve the performance of CBS and has become a common approach in leading MAPF ($k$ = 0) solvers.
%
%
%

%% file: symmetries.tex
\section{$k$-Symmetries}
\label{section:sym}

\begin{figure}
\centering
\begin{subfigure}{0.6\columnwidth}
  \centering
  \includegraphics[width=\columnwidth]{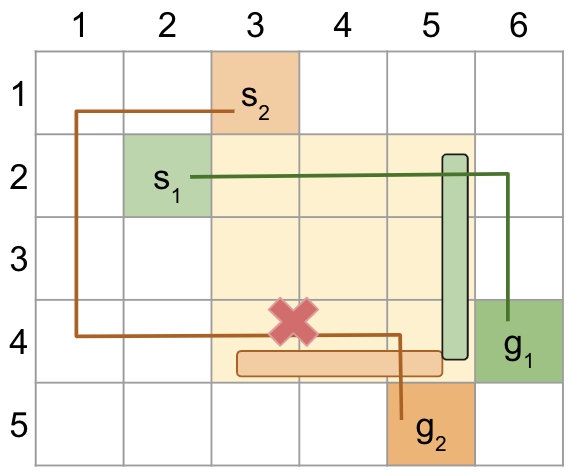}
    \caption{}
  \label{fig1:sub1}
\end{subfigure}
\begin{subfigure}{0.8\columnwidth}
  \centering
  \includegraphics[width=\columnwidth]{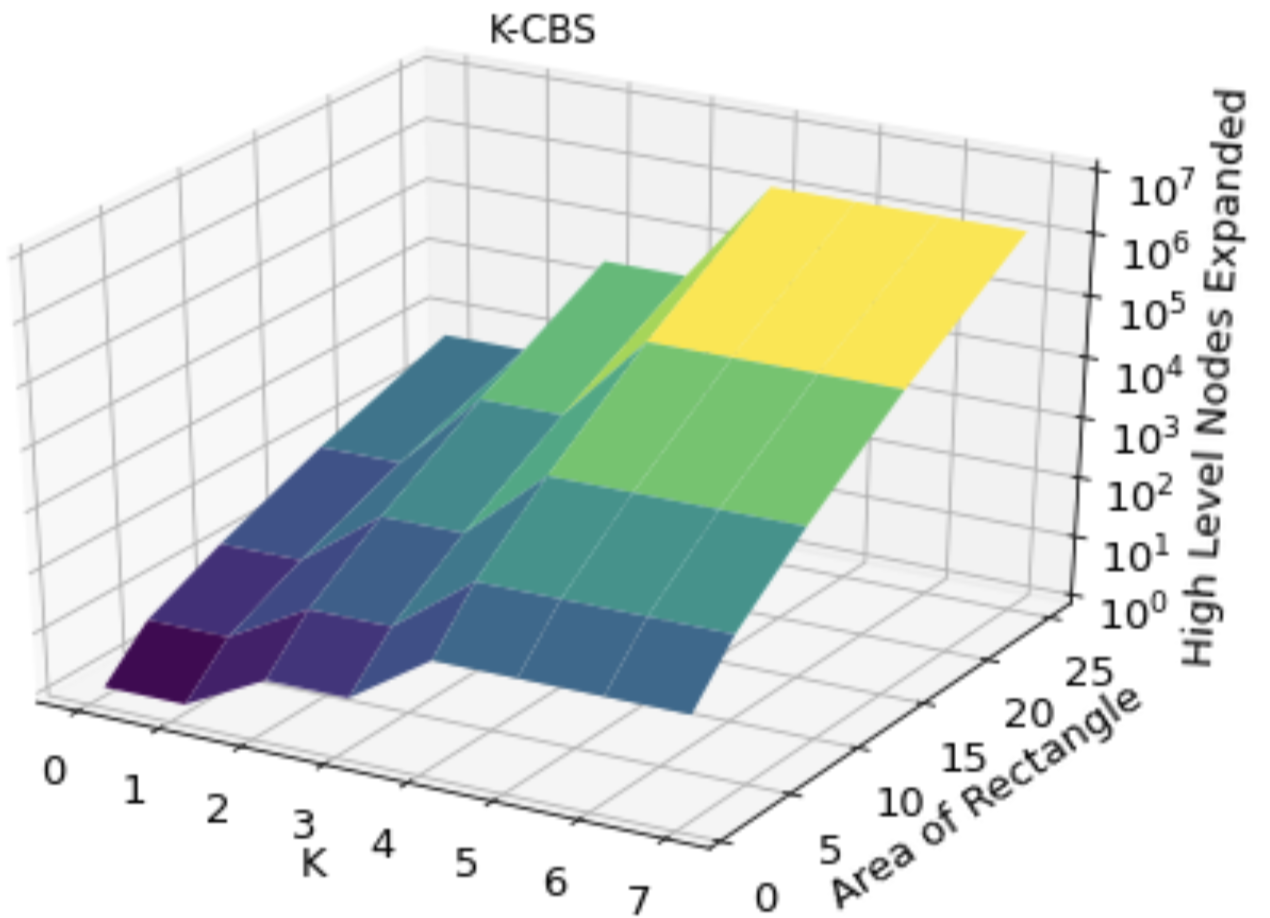}
    \caption{}
  \label{fig1:sub2}
\end{subfigure}
\caption{ (a)A collision-free solution, orange path, is eliminated, if we simply put "thick" barrier constraints to resolve a $k$-delay rectangle conflict when $k=4$.
(b) How number of high-level nodes expanded by $k$-CBS increases with $k$ and the size of the rectangle area. }
\label{fig1}
\end{figure}

Symmetries in  MAPF occur when two agents repeatedly run into one another
along equivalent individually optimal paths.  Figure~\ref{fig1:sub1} shows
an example for $k=4$.  Notice that each agent has available a number of optimal-cost
paths. However every optimal path of agent $a_1$ is in collision with
\emph{every} optimal path of agent $a_2$ and vice versa.
Without detecting such situations $k$-CBS will repeatedly split, node after
node, growing the CT tree in order to enumerate all possible
collisions in the highlighted rectangle area. Figure~\ref{fig1:sub2}
shows the result: each time $k$-CBS splits, it potentially doubles the 
amount of search yet remaining to reach the goal node. The size of the
CT tree grows exponentially as the size of the rectangle area increases, causing $k$-CBS 
to return timeout failure. 
Note that as $k$ increases, the size of the CT tree also grows exponentially,
which makes the problem extremely difficulty to solve. 
All this difficulty can be avoided by recognising there exists a simple
optimal strategy: one of the two agents has to wait. 
For $k = 0$ one agent must wait for one timestep. 
For $k > 0$ one agent may need to wait for more than one timestep. 
However, as $k$ grows large the problem can permit optimal-cost
bypass routes that allow one agent to avoid the rectangle area entirely.
For example in Figure~\ref{fig1:sub1}, when $k \geq 4$ the solid-orange-line 
path becomes optimal and agent $a_2$ can reach its target without waiting.

In this work we consider three
distinct symmetric situations which can appear in $k$-robust MAPF:
\begin{itemize}
\item Rectangle Symmetries, as illustrated in Figure~\ref{fig2:sub0}.
\item Corridor Symmetries, as illustrated in Figure~\ref{fig2:sub1}.
\item Target Symmetries, as illustrated in Figure~\ref{fig2:sub2}.
\end{itemize}

Rectangle symmetries have previously been studied in the context of MAPF
($k$ = 0)~\cite{li2019} while
Corridor and Target Symmetries have only recently been
introduced~\cite{corridor}, again in the context of MAPF ($k$ = 0).
Each time authors show that symmetries are common in a range of standard
benchmarks and they report dramatic gains in performance when these
symmetric situations are resolved via specialised reasoning techniques.
We adapt each of these ideas to $k$-robust planning and we report similarly
strong results. Generalising these constraints is not simply academic. As we
show in the experimental section, $k$-robust plans are needed for important
practical problems. Without the development of suitable algorithmic
techniques such problems will remain out of reach to MAPF planners.

%% file: reasoning.tex
\section{$k$-Robust Rectangle Reasoning}

Rectangle symmetries arise in $k$-CBS when the paths of two agents cross
topologically. The agents are heading in the same
directions (e.g. down and right in Figure~\ref{fig1:sub1}) and there
exists for each many different but equally shortest paths which arise
from re-ordering their individual moves.
In such cases the standard replanning strategy of $k$-CBS does not help to 
immediately resolve the problem. 

\begin{definition}[$k$-delay rectangle conflict]
A $k$-delay rectangle conflict between two agents occurs if all paths with cost between optimal and optimal+$k$ (inclusive) for the two agents that enter a given rectangular area have a $k$-delay vertex conflict in the rectangular area. 
\end{definition}

\noindent
Rectangle conflicts pose substantial challenges for $k$-robust planning
in general and for $k$-CBS in particular because:
\begin{itemize}
\item the agents do not have to reach positions at exactly the
same time to have a conflict; 
\item the delay caused by the conflict can
be up to $k+1$; and 
\item for $k\geq 2$ an agent can leave and enter the
conflicting area without adding more than $k$ steps to its path.
\end{itemize}
%
%
\noindent
To address these challenges we follow 
\citet{li2019} where authors develop {\em barrier constraints}:
a pruning strategy that can efficiently resolve rectangle conflicts 
for CBS with $k=0$ in a single 
branching step. With respect to Figure~\ref{fig1:sub1}, one barrier constraint 
prohibits agent $a_1$ from occupying cells (5, 2), (5, 3), and (5,4) at 
timesteps 3, 4, and 5, respectively. Similarly, the other barrier constraint prohibits agent $a_2$ 
from occupying cells (3, 4), (4, 4), and (5, 4) at timesteps 3, 4, and 5, respectively. 
Notice that each barrier blocks all equivalent shortest paths 
and forces one agent or the other to wait, thus resolving 
the conflict.

A straightforward idea for extending this strategy to $k > 0$ would be to increase the ``thickness'' (number of timesteps) of the barrier constraints. We therefore introduce \emph{temporal barrier constraints}, which unify the (temporal) range constraints of~\cite{atzmon} and the (spatial) barrier constraints of~\cite{li2019}.

\begin{definition}[Temporal barrier constraint]
Given a vertex-time pair $p = (u,t)$ we denote with $ot(p, v)$ the \emph{optimal time} 
to reach vertex $v = (v_x,v_y)$ from vertex $u = (u_x,u_y)$ starting at timestep $t$.
We compute the optimal time by adding to $t$ the Manhattan distance from $u$ to $v$:
\mbox{$ot(p,v) = t + |v_x - u_x| + |v_y - u_y|$}.
A \emph{$w$ temporal barrier constraint}, denoted $B(a_x, V, p, w)$, forbids an 
agent $a_x$, currently at vertex-time pair $p$, from visiting any vertex $v \in V$ 
at its optimal time or up to $w$ timesteps later. 
That is, the barrier constraint $B(a_x, V, p, w)$ 
is the set of time-range vertex constraints \mbox{$\langle a_x, v, [ot(p,v), ot(p,v) + w] \rangle, v \in V$}.
\label{defi:barrier}
\end{definition}

To resolve the $k$-robust rectangle conflict in Figure~\ref{fig1:sub1}, one might think that we can replace the two barrier constraints used by \cite{li2019} with two $k$ temporal barrier constraints at the exit of the rectangle. 
However, this approach is not complete! For example, when $k=4$, 
the orange line is a collision-free path for agent $a_2$, no matter what path agent $a_1$ takes. But this solution is eliminated by 
the $w=4$ temporal barrier constraints.
Therefore, we propose a novel approach that enlarges the rectangle area being reasoned about and adding temporal barrier constraints based on it with adjusted thickness so that we can preserve the completeness and optimality by taking into consideration such bypasses.

\subsection{Enlarging Rectangle and Shifting Borders}
\label{section:temporal}

\begin{figure}
\centering
\includegraphics[width=1\columnwidth]{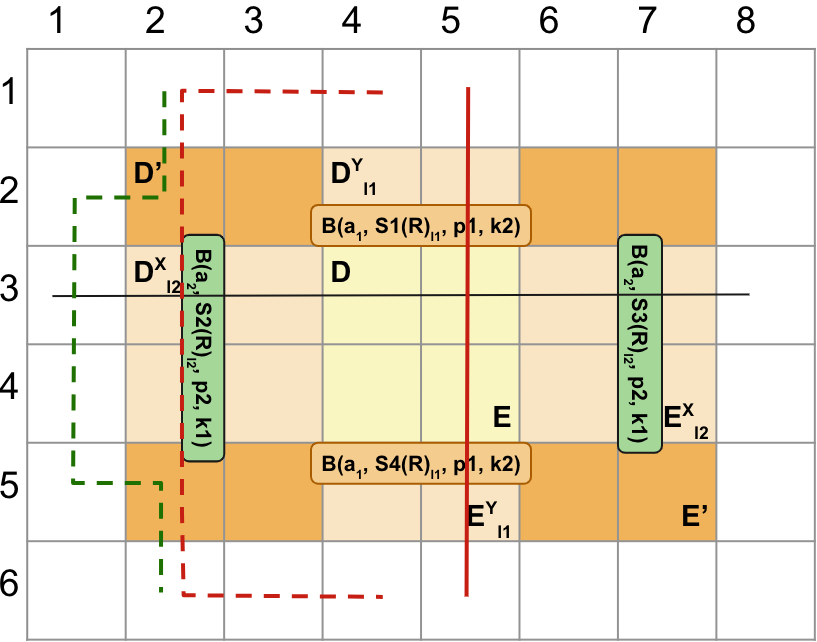}
\caption{(1) Example of temporal barrier constraints. In this case $k_1=2$, $S1(R)_{l_1}$ and $S4(R)_{l_1}$ are shifted for $l_1 = 1$ grid location away from $S1(R)$ and $S4(R)$,
and $k_2=4$ so $S2(R)_{l_2}$ and $S3(R)_{l_2}$ are shifted for $l_2 = 2$ timesteps from $S2(R)$ and $S3(R)$. (2) The yellow area is the rectangle $R$. Adding the light and dark orange areas
gives rectangle $R'$. (3) The solid red line is a path traversing rectangle $R$ and must have $k$-delay conflicts. The dashed red line is a path bypassing the rectangle $R$, but traversing rectangle $R'$ and must have $k$-delay conflicts as well. 
The green dashed line is a collision-free path bypassing the rectangle $R'$. (4) Extended constraints (Step Temporal Barrier Constraint) defined in Section \ref{section:stepTemporal} fill the gaps in the barriers for the dark orange area.
\label{fig-temporal}}
\end{figure}

Given a 4-neighbor grid map, we define a \emph{rooted rectangle}  $R$ as the set of vertices occurring in the rectangle defined by the 
two corner points $D=(D_x,D_y)$ the root corner, and 
$E=(E_x,E_y)$ the opposite corner. 
As shown in Figure \ref{fig-temporal}, the illustrated rectangle is  defined by $D=(4, 3)$ and $E=(5, 4)$. Given $D$ and $E$, we define four sides: 
$S1(R) = \{ (D_x,D_y) .. (E_x,D_y) \}$,
$S2(R) = \{ (D_x,D_y) .. (D_x,E_y) \}$,
$S3(R) = \{ (E_x,D_y) .. (E_x,E_y) \}$, and
$S4(R) = \{ (D_x,E_y) .. (E_x,E_y) \}$.
We define the \emph{shifted side}
$Sj(R)_l, j \in [1, 4], l \in [0, \lfloor \frac{k}{2}\rfloor]$,
as the side $Sj(R)$ shifted away from the center of $R$
by $l$ grid locations, note the $Sj(R)$ cannot be shifted beyond the start or goal locations of agents involved in the conflict.
Define the  $y$-shifted start location
$D^y_l$ as the root corner $D$ shifted $l$
locations along $y$-axis away from $R$,
and the $x$-shifted start location $D^x_l$ as the start location shifted $l$ locations along $y$-axis away from $R$. Similarly define $E^y_l$ and $E^x_l$.

Given these definitions, the shifted barrier constraints with some particular thickness can define a \emph{$k$-delay rectangle conflict}, that is all paths for two agents $a_1$ and
$a_2$ that cross these barriers must inevitably result in a $k$-delay conflict.

\begin{theorem}\label{thm:rect:1} 
Consider a $k$-robust MAPF problem with an arbitrary rooted rectangle $R$ defined by corners $D$ and $E$
and two arbitrary integers $0 \leq k_1,k_2 \leq k$.
Let root time $rt$ be the minimum of the timestep when agent $a_1$ or $a_2$ reaches the root corner $D$, $l_1 = \lfloor \frac{k_1}{2} \rfloor$,  $l_2 = \lfloor \frac{k_2}{2} \rfloor$, 
$p_1 = (D^y_{l_1},rt-l_1)$, and $p_2 = (D^x_{l_2},rt-l_2)$.
If the paths for agents $a_1$ and $a_2$ violate all of the four constraints:
\begin{itemize}
    \item \emph{$a_1$ Entrance:} $B(a_1, S1(R)_{l_1}, p_1, k_2)$,
    \item \emph{$a_1$ Exit:}     $B(a_1, S4(R)_{l_1}, p_1, k_2)$,
    \item \emph{$a_2$ Entrance:} $B(a_2, S2(R)_{l_2}, p_2, k_1)$,
    \item \emph{$a_2$ Exit:}     $B(a_2, S3(R)_{l_2}, p_2, k_1)$,
\end{itemize}
\noindent 
the paths of the two agents have a $k$-delay vertex conflict. 
\end{theorem}
\begin{proof}
Assume that $S1(R)$ is the top of the rooted rectangle $R$, 
$S4(R)$ the bottom, $S2(R)$ the left, and $S3(R)$ the right. The other cases follow similarly.
$S1(R)_{l_1}$, $S4(R)_{l_1}$, $S2(R)_{l_2}$, $S3(R)_{l_2}$ are corresponding sides shifted away from $R$.
These shifted sides define a new larger rectangle $R'$ with two corner points $D'$ and $E'$, as shown in Figure~\ref{fig-temporal}.

In order to violate the first two constraints, agent $a_1$ has to enter the top of rectangle $R'$ through a vertex $v$ on $S1(R)_{l_1}$ at timestep $ot(p_1,v)$ or up to $k_2$ timesteps later, and leave from the bottom through a vertex $v$ on $S4(R)_{l_1}$ at timestep $ot(p_1,v)$ or up to $k_2$ timesteps later,
so it can wait for at most $k_2$ timesteps in $R'$ but cannot leave $R'$, since this would require at least extra $2l_2 + 2$ timesteps comparing with a shortest path across $R'$ but $2l_2 + 2 > k_2$.
Every vertex $v$ it visits in $R'$ is visited 
within the time range [$ot(p_1,v), ot(p_1,v) +k_2]$.

Similarly agent $a_2$ enters the left of rectangle $R'$ through a vertex $v$ on $S2(R)_{l_2}$ at timestep $ot(p_2,v)$ or up to $k_1$ timesteps later, and leaves the right through a vertex $v$ on $S3(R)_{l_2}$ at timestep $ot(p_2,v)$ or up to $k_1$ timesteps later. 
Again it cannot leave $R'$, since this would require at least extra $2l_1 + 2$ timesteps and it can take at most $k_1$ wait in $R'$. Every vertex $v$ it visits in $R'$ is visited
within the time
range $[ot(p_2,v), ot(p_2,v) + k_1]$.

Now since agent $a_1$ crosses from top to bottom and agent $a_2$ from left to right, their paths must cross, say at vertex $v \in R'$. Assume that agent $a_1$ visits vertex $v$ at $t_1 \in [ot(p_1,v), ot(p_1,v) + k_2]$ while agent $a_2$ visits vertex $v$ at $t_2 \in [ot(p_2,v), ot(p_2,v) + k_1]$.
Since $ot(p_1,v)$ and $ot(p_2,v)$ share the same root time $rt$, $ot(p_1,v) = ot(p_2,v)$, and $k_1$ and $k_2$ are both less than or equal to $k$,
which make $|t_1 - t_2| \leq k$,
there is a $k$-delay conflict
$\langle a_1, a_2, v, t_1, t_2-t_1 \rangle$ (if $t_1 \leq t_2$) or $\langle a_2, a_1, v, t_2, t_1-t_2 \rangle$ (if $t_1 > t_2$).
\end{proof}

\subsection{Extending Temporal Barrier Constraints}
\label{section:stepTemporal}

The temporal barrier constraints proposed in Theorem 1 do not cover all situations where two agents must have a $k$-delay vertex conflict in a rectangle area because they do not cover the entire circumference of rectangle $R'$. As the red dashed line path shown in Figure \ref{fig-temporal} illustrates, if agent $a_1$ enters the top of rectangle $R'$ through corner $D'$ at the optimal time, and leaves from left-bottom vertex at the optimal time. This path (red dashed line) traverses through $S2(R)_{l_2}$ and is $2\lfloor \frac{k}{2} \rfloor$ timesteps longer than paths entering through  $S1(R)_{l_1}$ optimally and leaving through $S4(R)_{l_1}$ optimally. Now agent $a_2$ enters $R'$ via $S2(R)_{l_2}$ at the optimal time or up to $k_2$ timesteps later, so they still have a $k$-delay conflict.

We cannot simply extend temporal barrier constraints to the entire circumference of $R'$ to eliminate such conflicts, 
as it would also eliminate the green dashed collision-free path in Figure \ref{fig-temporal}. 
We thus consider a stronger reasoning method.

\begin{definition}[Step temporal barrier constraint]
Denoted by $B_{step}(a_x, S, p, k')$ this constraint reduces the temporal width of 
the barrier $B(a_x, S, p, k')$ by a value of 2 for each step away from the original 
barrier. Let $l' = \lfloor \frac{k'}{2} \rfloor$, we add
range constraints 
$\langle a_x, v, [ot(p,v), ot(p,v)+k' - 2d] \rangle$
for each vertex $v$ in the same line as $S$ 
at distance $d \in [1, l']$ from the original barrier.
\end{definition}

\ignore{
To eliminate these kind of conflicts. We introduce \textbf{Step Temporal Barrier Constraints} denoted by $B_{step}(a_x, V, V_{root}, p, k)$ which is a set of range constraints $\langle a_x, v_i, range(v_i) \rangle, v_i \in V \And v_i\notin V_{root}$.

\pjs{FIX!}
$V$ and $V_{root}$ are two sets of vertexes. Agent $a_x$ is forbid to visit $v_i$, where $v_i \in V$ and $v_i \notin V_{root}$, for a time range. This time range is derived by function $range(v_i)$:
\[
    range(v_i)= 
\begin{cases}
    [ot(p,v_i), upper(v_i)],& \text{if } upper(v_i) \geq ot(p,v_i)\\
    None,              & \text{otherwise}
\end{cases}
\]
\[
upper(v_i) = ot(p,v_i) + k - 2 \cdot d(v_i,V_{root}) 
\]
\[
d(v_i,V_{root}) = \min_{v_j\in V_{root}}Manhattan Distance(v_i, v_j) 
\]
}

For the rectangle $R'$ in Figure~\ref{fig-temporal}, the step temporal barrier constraint
$B_{step}(a_1, S1(R)_{l_1},p_1, k_2)$
consists of $B(a_1, S1(R)_{l_1}, p_1, k_2)$
and the additional range constraints
$\langle a_1, (2,2), [rt+1, rt+1]\rangle$,
$\langle a_1, (3,2), [rt, rt+2]\rangle$,
$\langle a_1, (6,2), [rt+1, rt+3]\rangle$, and
$\langle a_1, (7,2), [rt+2, rt+2]\rangle$.
Notice how the time ranges shrink further from the original barrier.
This prevents the red dashed path in Figure~\ref{fig-temporal}.
We can now extend Theorem~\ref{thm:rect:1}: if the paths of agents $a_1$ and $a_2$ violate all of the four following constraints: 
\begin{itemize}
    \item \emph{$a_1$ Entrance:} $B_{step}(a_1, S1(R)_{l_1},p_1, k_2)$
    \item \emph{$a_1$ Exit:} $B_{step}(a_1, S4(R)_{l_1},p_1, k_2)$
    \item \emph{$a_2$ Entrance:}  $B_{step}(a_2, S2(R)_{l_2},p_2, k_1)$ 
    \item \emph{$a_2$ Exit:} $B_{step}(a_2, S3(R)_{l_2},p_2, k_1)$
\end{itemize}
the paths of the two agents have a
$k$-delay vertex conflict. The proof is similar.

\subsection{Resolution of $k$-Delay Rectangle Conflicts}

We can always 
resolve a rectangle conflict by four-way branching adding to each branch one of the constraints:
$a_1$ Entrance, $a_1$ Exit, 
$a_2$ Entrance, and $a_2$ Exit;
since we know that one of them must be violated in any solution.
\ignore{
$$B_{step}(a_x, S1(R'),S1(R)_{l_1},p_1, k_2) \wedge B(a_1, S1(R)_{l_1},p_1, k_2)$$
$$\vee$$ 
$$ B_{step}(a_x, S4(R'),S4(R)_{l_1},p_1, k_2) \wedge B(a_1, S4(R)_{l_1},p_1 , k_2)$$
$$\vee$$ 
$$ B_{step}(a_x, S2(R'),S2(R)_{l_2},p_2, k_1) \wedge B(a_2, S2(R)_{l_2}, p_2, k_1) $$
$$\vee$$ 
$$B_{step}(a_x, S3(R'),S3(R)_{l_2},p_2, k_1) \wedge B(a_2, S3(R)_{l_2},p_2 , k_1) $$
}
But in many cases, we can correctly resolve the rectangle conflict by two-way branching
on the constraints  $a_1$ Exit and $a_2$ Exit,
if  both agents satisfy the following condition:
\begin{enumerate}[label=\bfseries Condition \arabic*,wide]
\item All possible paths that traverse the exit barrier must also traverse the entrance barrier.\label{condition:1}
\end{enumerate}

We can use a $k$-MDD to check that the condition is satisfied. A $k$-MDD for agent $a_i$ is a modified Multi-Valued Decision Diagram (MDD)~\cite{boyarski2015} that stores all paths of agent $a_i$ from start to goal with path length no more than $k$ above the optimal. 
MDDs are widely used in CBS algorithms to store all optimal paths, the $k$-MDD is a direct extension.

If either agent's $k$-MDD shows that paths that bypass the entrance barrier and traverse the exit barrier exist, the given conflict cannot be resolved by two-way branching on the given barriers, as it may eliminate conflict free paths that bypass the entrance barrier. 
If any combination of $k_1$ and $k_2$ lead to
\ref{condition:1} being satisfied, the given conflict can resolved by two-way branching on the exit barriers.
If none of the combinations satisfy \ref{condition:1}, the given conflict will be resolved as a normal conflict
(i.e. we never do four-way branching).

We can classify $k$-delay rectangle conflicts as cardinal, semi-cardinal and non-cardinal using the $k$-MDD:
\begin{itemize}
    \item A $k$-delay rectangle conflict is \emph{cardinal}, if all paths in the $k$-MDDs of both agents traverse the exit barrier, which means that replanning for any agent involved in the conflict increases the SIC.
    \item A $k$-delay rectangle conflict is \emph{semi-cardinal}, if only one agent has all paths in its $k$-MDD traverse the exit barrier, which again means that replanning this agent involved always increases the SIC while replanning for the other agent does not.
    \item A $k$-delay rectangle conflict is \emph{non-cardinal} if both agents have paths in their $k$-MDD bypass their exit barriers.
\end{itemize}
We can then prioritize selecting conflict based on cardinality.

\subsection{Detecting Rectangle Conflicts}
\label{section:detectRect}

When we detect a vertex conflict during CBS we need to recognise that it's actually a 
rectangle conflict, in order to perform rectangle symmetry breaking.

Assume that agents $a_1$ and $a_2$ have a $k$-delay vertex conflict $\langle a_1, a_2, v, t, \Delta \rangle$. Let $d_1$ and $d_2$ be the moving directions when agents $a_1$ and $a_2$ enter vertex $v$, respectively. If they are the same directions, then there is an earlier vertex conflict (where they both came from). 
If they are opposite directions, then there is no rectangle conflict.

So assume $d_1$ and $d_2$ are orthogonal directions.
Let $(B1_x,B1_y)$ be the earliest vertex in the path of agent $a_1$ where all moves from here to $v$ are in direction $d_1$ or $d_2$, similarly define $(B2_x,B2_y)$ for agent $a_2$. Let $t_{b1}$ be the earliest timestep that $a_1$ visits $(B1_x,B1_y)$ and $t_{b2}$ be the earliest timestep that $a_2$ visits $(B2_x,B2_y)$.
Let $(A1_x,A1_y)$ be the latest vertex in the path of agent $a_1$ where all moves from $v$ to here are in directions $d_1$ or $d_2$, similarly define $(A2_x,A2_y)$ for agent $a_2$.

\begin{figure}
  \centering
  \includegraphics[width=0.7\columnwidth]{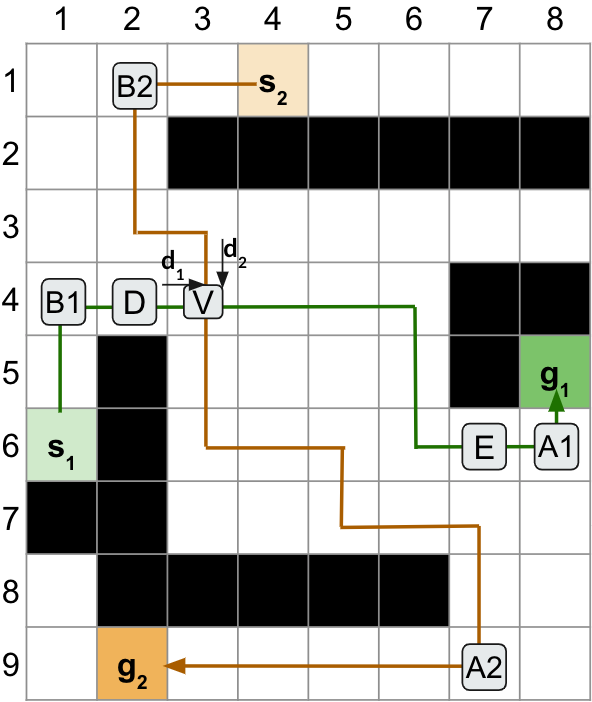}
  \caption{Detecting a rectangle conflict when $k=2$. }
 
  \label{RecDectFig}
\end{figure}

Define $D_x$ to be the closer of $B1_x$ and $B2_x$ to $v_x$,
similarly for $D_y$. Define $E_x$ to be the closer of $A1_x$
and $A2_x$ to $v_x$, similarly for $E_y$. Figure \ref{RecDectFig} shows an example.

Let $rt_1 = t_{b1} + |B1_x - D_x| + |B1_y - D_y|$ and $rt_2 = t_{b2} + |B2_x - D_x| + |B2_y - D_y|$. 
We define the root time $rt = \min (rt_1,rt_2)$. 

We then for each value $k_1 \in \{0,1,\ldots,k\}$
and $k_2 \in \{0,1,\ldots,k\}$ check if the agents
satisfy \ref{condition:1} using the 
Step Temporal Barriers defined by these values.
We do so by examining the $k$-MDD for each agent,
temporarily blocking its entrance barrier and seeing if its exit barrier is still reachable. 
If not then \ref{condition:1} holds. 
We try the values for $k_1$ and $k_2$ in decreasing order to find the strongest blocking conditions possible.
If $k_1 = a, k_2 = b$ satisfies the conditions we don't
investigate any pairs $(a',b')$ where $a' \leq a$ and 
$b' \leq b$.

Figure \ref{RecDectFig} shows an example of the rectangle detection. In this example,  agents $a_1$ and $a_2$ have a 2-delay conflict $\langle a_1,a_2,(3,4),4,2\rangle$ with $d_1$ pointing right and $d_2$ pointing down. Then $B_1$ is located at $(1,4)$ with $t_{b1} = 2$ and $rt_1 = 3$, and $B_2$ located at $(2,1)$ with $t_{b2} = 2$ and $rt_2 = 5$. Root time $rt = min(t_{b1}, t_{b2}) = 3$. 
We detect that for the case $k_1 = k_2 = 2$ the 
Step Temporal Barriers defined by these values satisfy \ref{condition:1}.

We call the detection of rectangles and the associated branching K-CBSH-RM (it is a generalization of the RM technique defined by~\citet{li2019}). 

\begin{theorem}
\label{thm:rect:2}
K-CBSH-RM is correct.
\begin{proof}
Since \ref{condition:1} holds for the chosen 
Step Temporal Barriers, and using (extended) Theorem \ref{thm:rect:1}, each pair of paths that violate the exit barriers must conflict. Hence the two-way branching removes no solutions, and advances the search since it is violated by the current paths.
\end{proof}

\end{theorem}

%% file: K-Robust Symmetry Breaking (4)/K_Corridor_Target.tex
\section{$k$-Robust Corridor Reasoning}

A \emph{corridor} from $B$ to $E$ is a chain of nodes $C$ where nodes in $C$ except $B$ and $E$
each have exactly two neighbours, 
and $B$ and $E$ have exactly one neighbour in $C$.
Figure~\ref{fig2:sub1} shows a corridor where $B = (3, 1)$, $C = \{ (1, 3), (2, 3),(3,3), (4, 3), (5, 3) \}$ and $E = (5, 3)$.
A \emph{corridor conflict}~\cite{corridor} 
occurs when two agents have a vertex or edge conflict occurring in a corridor.
In the example agents $a_1$ and $a_2$ conflict at vertex
(3, 3). 
Simply adding a vertex conflict constraint will not resolve the conflict, they will continue to
conflict in the corridor. \citet{corridor} introduce corridor symmetry breaking constraints which we extend here for $k$-delay conflicts.

The difference between corridor conflicts for $k$-robust CBS and normal corridor conflicts is that agents need to occupy vertexes for extra timesteps to avoid $k$-delay conflicts. Assuming there is a corridor with length of $l$ between vertex $B$ and vertex $E$, and a $k$-delay conflict in the corridor with agent $a_1$ ($a_2$) moving from $B$ to $E$ (resp. $E$ to $B$). Let $t_1$ (resp. $t_2$) be the earliest timestep when agent $a_1$ ($a_2$) is able to reach vertex $E$ (resp. $B$). 

Clearly when planning a $k$-robust solution, any path of $a_1$ using the corridor that reaches vertex $E$ at or before timestep $t_2+l+k$ must conflict with any paths of $a_2$ using the corridor that reach vertex $B$ at or before timestep $t_1+l+k$. But there may be alternate paths the agents can take to reach $B$ or $E$.
Assuming agent $a_1$ can reach vertex $E$ at timestep $t'_1$ without using the corridor 
and reach vertex $B$ at timestep $t_b$. Agent $a_2$ can reach vertex $B$ at timestep $t'_2$ without traversing the corridor and reach vertex $E$ at timestep $t_e$. 
Hence in planning a $k$-robust solution, any path of $a_1$ that reaches vertex $E$ at or before timestep $min( max(t_e + k, t'_1 - 1), t_2  + l+k)$ must conflict with any path of $a_2$ that reaches vertex $B$ at or before timestep $min( max(t_b + k, t'_2  - 1),t_1  + l+k)$.

Hence the constraint $\langle a_1,E,[0,min( max(t_e + k, t'_1 - 1), t_2  + l+k)] \rangle \vee \langle a_2,B,[0,min( max(t_b + k, t'_2  - 1),t_1  + l+k)] \rangle$ must hold in all solutions. 
To handle the corridor constraint we branch on this disjunction. Clearly

\begin{theorem}$k$-robust Corridor Reasoning is correct.
\qed
\end{theorem}
\ignore{
\begin{proof}
We have shown that if $a_1$ and $a_2$ both violate constraint $<a_1,E,[0,min(t'_1  - 1+k,t_2  + l+k)]>$ and constraint $<a_2,B,[0,min(t'_2  - 1+k,t_1  + l+k)]>$, they must have a conflict in the corridor. Thus the adding of these constraints will not eliminate any conflict free $k$-robust solution, and will eliminate the current paths.
\end{proof}
}

\section{$k$-Robust Target Reasoning}

A \emph{target conflict}~\cite{corridor} occurs when one agent $a_2$ reaches its goal vertex $g_2$ at timestep $l$, and another agent $a_1$ conflicts with agent $a_2$ at vertex $g_2$ at some later timestep 
$t, t \geq l$.  
Consider Figure~\ref{fig2:sub2} where agent $a_2$ reaches its goal cell (4,2) at timestep 1, and then agent $a_1$ tries to traverse cell (4,2) at timestep 3.  Simply adding the constraint
$\langle a_1,(4,3),[3,3] \rangle$ causes $a_1$ to wait before entering 
cell (4,3) at timestep 4 and then the conflict reoccurs.

To avoid this ~\citet{corridor} resolve the conflict by branching on $et_2 \leq t \vee et_2 > t$
where $et_2$ is the end time for agent $a_2$. In the first case, since agent $a_2$ finishs before or at timestep $t$, agent $a_1$ can never use location $g_2$ at timestep $t$ or after. In the second case agent $a_2$ cannot finish before timestep $t+1$ freeing up the location for agent $a_1$. 

Planning a $k$-robust solution requires the avoiding of $k$-delay conflict. 
Therefore, we branch on $et_2 \leq t+k \vee et_2 > t + k$. The first case forces agent $a_2$ to finish
before timestep $t+k$ preventing agent $a_1$ (or any other agent) from using vertex $g_2$ at timestep $t$, the second case
forces agent $a_2$ not to finish earlier so vertex $g_2$ at timestep $t$ is freed up for agent $a_1$. Again clearly

\begin{theorem}
$k$-Robust Target Reasoning is correct.
\ignore{
\begin{proof}
The disjunction we branch on clearly removes no solutions, and since each disjunct makes one of current paths infeasible, it progresses the search.
\end{proof}}
\qed
\end{theorem}

Note that to handle target symmetries we have to update the low-level path finder for agents to take into account new kinds of constraints where we restrict the end time of an agent, and where we prevent any agent from using a location from some time point onwards. Both are straightforward additions. See~\citet{corridor} for details.

%% file: figures.tex
\begin{figure*}[t]
\centering
\begin{subfigure}{1\textwidth}
  \centering
  \includegraphics[width=0.8\columnwidth]{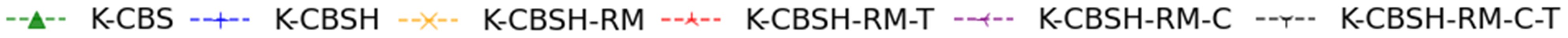}
\end{subfigure}
\begin{subfigure}{0.16\textwidth}
  \centering
  \includegraphics[width=\columnwidth]{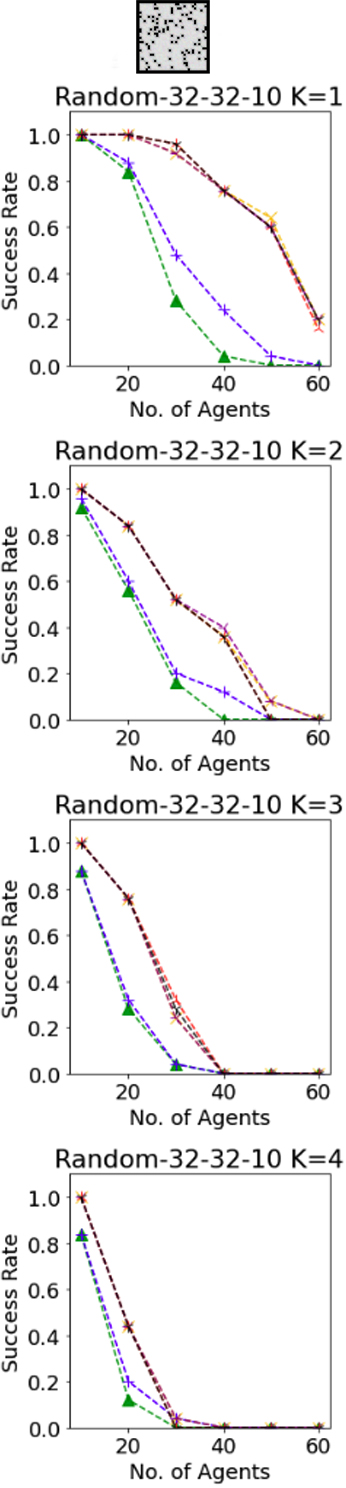}
  \caption{Random 32x32}
  \label{fig7:sub1}
\end{subfigure}
\begin{subfigure}{0.16\textwidth}
  \centering
  \includegraphics[width=\columnwidth]{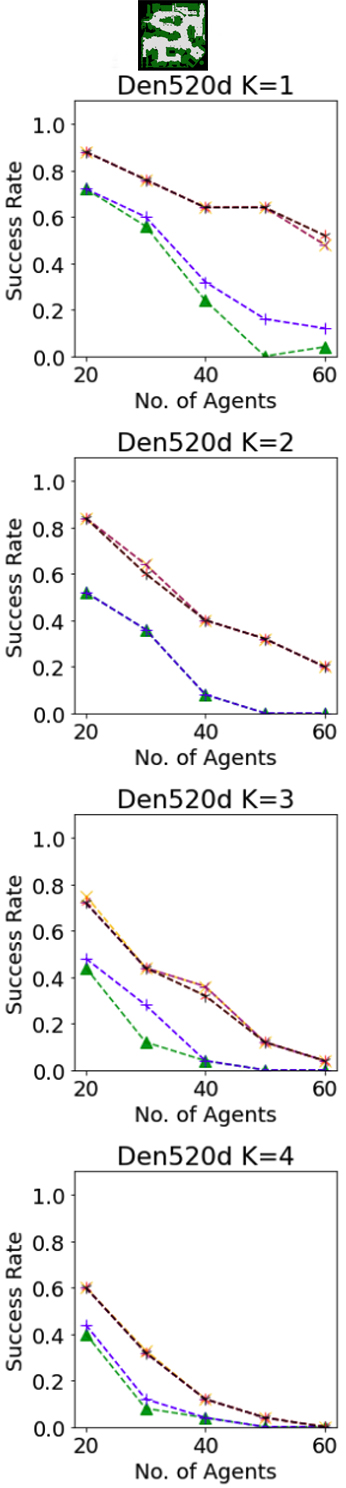}
  \caption{Den520d}
  \label{fig7:sub2}
\end{subfigure}
\begin{subfigure}{0.16\textwidth}
  \centering
  \includegraphics[width=\columnwidth]{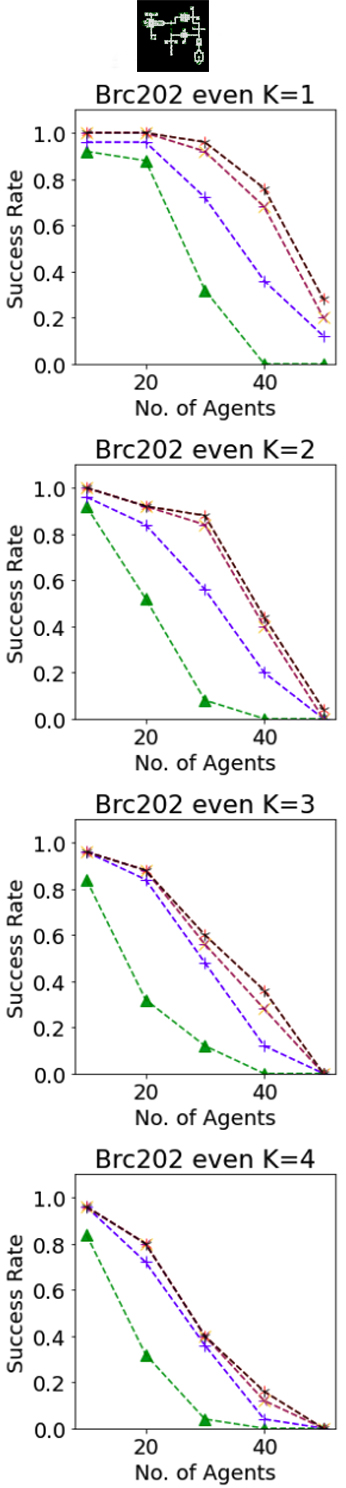}
  \caption{Brc202d}
  \label{fig7:sub3}
\end{subfigure}
\begin{subfigure}{0.16\textwidth}
  \centering
  \includegraphics[width=\columnwidth]{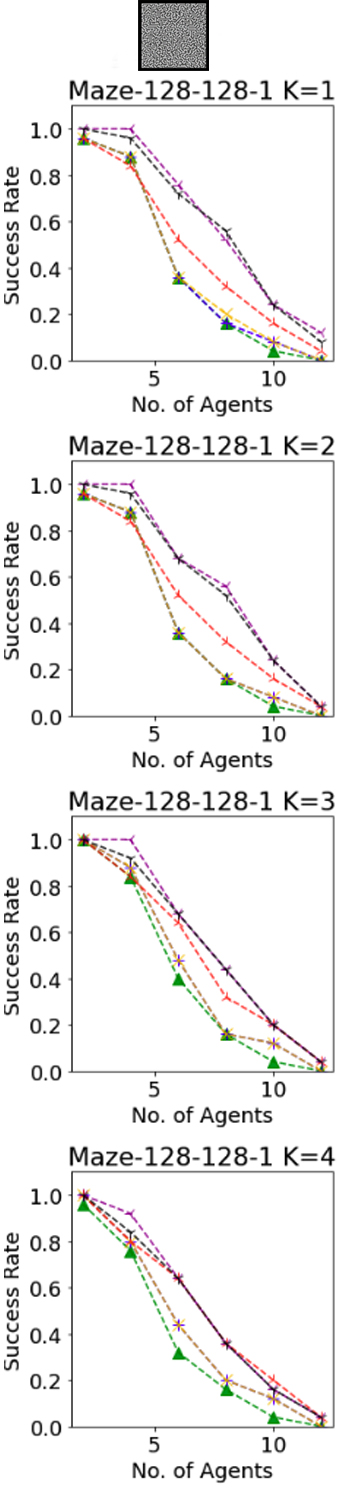}
  \caption{Maze 128x128}
  \label{fig7:sub4}
\end{subfigure}
\begin{subfigure}{0.16\textwidth}
  \centering
  \includegraphics[width=\columnwidth]{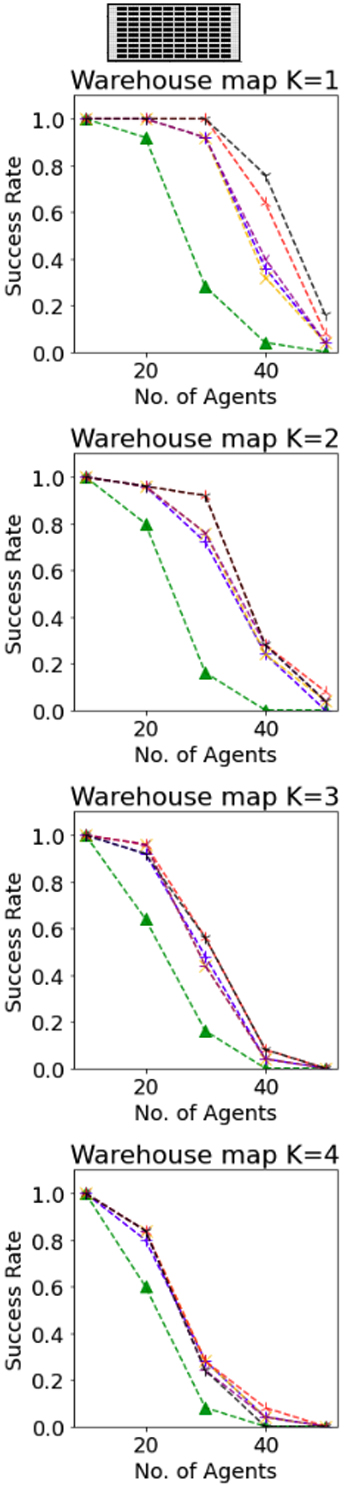}
  \caption{Warehouse map}
  \label{fig7:sub5}
\end{subfigure}
\begin{subfigure}{0.16\textwidth}
  \centering
  \includegraphics[width=\columnwidth]{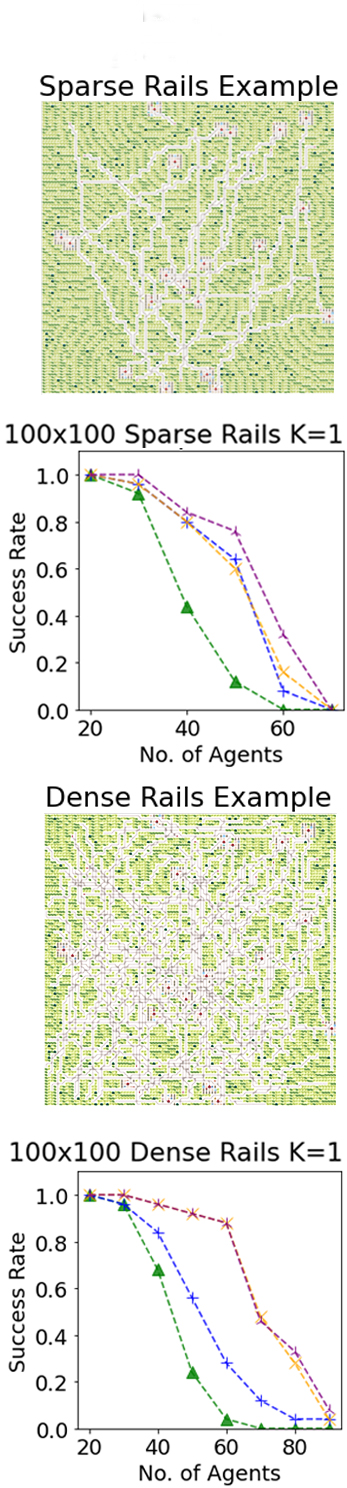}
  \caption{Flatland Railways}
  \label{fig7:sub6}
\end{subfigure}
\caption{Success rate versus number of agents on different problems.}
\label{fig7}
\end{figure*}

%% file: experiments.tex
\section{Experiments}

The implementation is based on CBS with rectangle, corridor and target reasoning of \citet{corridor} and support for $k$-robust planning is added on top of it. It is programmed in C++ and experiments were performed on a server with AMD Opteron 63xx class CPU and 32 GB RAM. For each map, we keep increasing the number of agents, and for each number of agent we solve 25 different instances. The time limit is set to 90s for each instance. In result plots, K-CBS is the current state of art algorithm to plan $k$-robust plan proposed by \citet{atzmon},which is selected as the baseline, K-CBSH 
is our extension of K-CBS with heuristics (Section~\ref{sec:kCBS}),
RM adds $k$-robust Rectangle reasoning, 
C adds $k$-robust Corridor Reasoning, 
and T adds $k$-robust Target Reasoning.

\subsubsection{Experiment 1: Game Maps}
The MAPF research community have developed a series of benchmark maps from games \cite{Stern2019}. They are available from movingai.com. We use 25 even scenarios from movingai.com to evaluate our algorithms.
We run experiments on following representative maps: \emph{Brc202d}, \emph{Den520d}, \emph{Random-32-32-10}, and \emph{Maze-128-128-1}. 

Figure \ref{fig7} shows the experiment results on grid game maps. K-CBSH-RM based algorithms shows significant higher success rate compared with K-CBS on \emph{Brc202d}, \emph{Den520d}, and \emph{Random-32-32-10}. Although $k$-robust corridor reasoning does not help on these three maps and $k$-robust target reasoning slightly helps, they effectively improves the success rate on \emph{maze-128-128-1}.
As $k$ increases, the problem becomes harder, and the success rate drops, but the symmetry-breaking algorithms still show significant advantages over K-CBS. 

\subsubsection{Experiment 2: Warehouse Map}
We use a \emph{31$\times$79 Warehouse map} \cite{corridor} with randomly generated problems to evaluate the performance of robots in warehouse system.
Figure \ref{fig7:sub5} shows that k-CBSH significantly improves success rate, $k$-robust target reasoning and corridor reasoning helps  to further improve the success rate of K-CBSH-RM.
Here target reasoning is clearly very important.

\subsubsection{Experiment 3: Simplified Railway System }

The Flatland challenge is a railway scheduling challenge \footnote{Swiss Federal Railways, 2019, Flatland Challenge. https://www.aicrowd.com/challenges/flatland-challenge} provides a simplified railway simulator using a directed grid map, where trains cannot move backwards. Railway systems have headway control, one train cannot start to enter a railway block if another train currently occupies
the block. Hence the railway domain requires $k=1$ robust plans.
Our experiments use flatland-rl v2.0.0 to generate experiment problems. 
Note, no target conflicts 
can occur in the Flatland challenge
scenarios since trains ``disappear'' when reaching their destination.

We have two settings for evaluation: (1) a \emph{100$\times$100 fixed dense map} contains fixed 140 start and goal locations, and each pair of start and goal are connected by 5 railways; and (2) a \emph{100$\times$100 incremental sparse map} has one start and goal location per agent, and each pair of start and goal are only connected by 1 railway.

The experiment on railway maps, Figure \ref{fig7:sub6} shows that K-CBSH and K-CBSH-RM performs substantially better than K-CBS. $k$-robust corridor reasoning helps further improve performance on sparse railways map.
Clearly the rectangle methods are more important on the denser map, where more symmetric conflicts are possible, and $k$-robust corridor reasoning is more important on sparse maps.

\subsubsection{Ratio of Rectangle Conflicts}

\begin{figure}
\centering
\includegraphics[width=0.8\columnwidth]{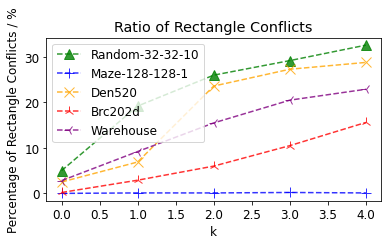}
\caption{Percentage of $k$-delay rectangle conflicts among all conflicts as $k$ increases when using K-CBSH-RM-C-T.}
\label{fig-rec-ratio}
\end{figure}

The Figure \ref{fig-rec-ratio} shows the percentage of $k$-delay rectangle conflicts among all resolved conflicts using K-CBSH-RM-C-T as $k$ increases. The statistics on conflicts are derived from Experiment 1 and Experiment 2. 
Clearly, as $k$ increases, the percentage of $k$-delay rectangle conflicts rises on the maps 
where rectangle conflicts can occur, hence the 
importance of $k$-robust rectangle reasoning is demonstrated. 

%% file: conclusion.tex
\section{Conclusions and Future Work}
This research introduces symmetry resolution methods for generating $k$-robust plans, which are vital for 
robust (e.g. warehouse robotics) and safe (e.g. railway scheduling) multi-agent plans. 
%
Symmetry reasoning methods improve dramatically on K-CBS, with $k$-robust rectangle reasoning being the most important, 
while $k$-robust 
corridor 
and target reasoning can further improve the performance. 
\ignore{
Future work crucially involves resolving the symmetric conflicts that can occur for $k\geq 2$  when two agents moving in the same direction in one dimension and the opposite direction in another dimension conflict. These conflicts 
occur fairly frequently and cause problems for $k$-CBS.
}

\section{Acknowledgments}

Research at Monash University was partially supported by the Australian Research Council under grants DP190100013, DP200100025 and also by a gift from Amazon.